\documentclass[jou]{apa6}

\usepackage[ruled]{algorithm2e}

\SetAlFnt{\small}
\SetAlCapFnt{\small}
\SetAlCapNameFnt{\small}
\SetAlCapHSkip{0pt}
\IncMargin{-\parindent}

\usepackage{lineno,hyperref}
\modulolinenumbers[5]

\usepackage{amssymb}
\usepackage{graphicx}
\usepackage[retainorgcmds]{IEEEtrantools}
\usepackage[cmex10]{amsmath}
\usepackage{float}
\usepackage{multirow}
\usepackage{multicol}
\DeclareMathOperator*{\argmax}{arg\,max}

\usepackage{url}
\usepackage{color}

\usepackage{chngcntr}

	\threeauthors{Naveen Nair}{Ajay Nagesh}{Ganesh Ramakrishnan}
\threeaffiliations{Department of Computer Science and Engineering\\ IIT Bombay, India \\ \texttt{naveennair@cse.iitb.ac.in}}{Department of Computer Science and Engineering\\ IIT Bombay, India \\ \texttt{ajaynagesh@cse.iitb.ac.in}}{Department of Computer Science and Engineering \\ IIT Bombay, India \\ \texttt{ganesh@cse.iitb.ac.in}}

    \title{Learning Discriminative Relational Features for Sequence Labeling}
    \shorttitle{Learning  Discriminative Relational Features for Sequence Labeling}
\abstract {Discovering relational structure between input features in sequence labeling models has shown to improve their accuracy in several problem settings. However, the search space of relational features is exponential in the number of basic input features. Consequently, approaches that learn relational features, tend to follow a greedy search strategy. In this paper, we study the possibility of optimally learning and applying discriminative relational features for sequence labeling. For learning features derived from inputs at a particular sequence position, we propose a Hierarchical Kernels-based approach (referred to as Hierarchical Kernel Learning for Structured Output Spaces - StructHKL). This approach optimally and efficiently explores the hierarchical structure of the feature space for problems with structured output spaces such as sequence labeling. Since the StructHKL approach has limitations in learning complex relational features derived from inputs at relative positions, we propose two solutions to learn relational features namely, (i) enumerating simple component features of complex relational features and discovering their compositions using StructHKL and (ii) leveraging relational kernels, that compute the similarity between instances implicitly, in the sequence labeling problem. We perform extensive empirical evaluation on publicly available datasets and record our observations on settings in which certain approaches are effective.
}

\begin{document}
	
\maketitle

\section{Introduction}
\label{sec:intro}

Structured output\footnote{ Output here is referring to the target or response, which is commonly referred to as output label in classification problems. Here our output is not a single variable but a structure constructed using multiple labels.}
classification has gathered significant interest in the machine-learning community during the last decade \cite{TsochantaridisSVMStructured, structuredPrediction5, structuredPrediction6}. The goal of such works is to classify complex output structures such as sequences, trees, lattices or graphs, in which the class label at each node/position of the structure has to be inferred from other observable variables.  The possible space of structured outputs tends to be exponential, and thus, structured output classification is a challenging research task. In our research work, we focus on a specific structured output classification problem, popularly known as sequence labeling. Activity recognition, named entity recognition, part-of-speech tagging, {\it etc.} are a few of the application areas of sequence labeling. In this paper, we use activity recognition domain for illustration and evaluation of our approaches. In activity recognition settings, the objective is to automatically label each time instance with the activity performed by a user, based on readings from sensors fitted at different locations in the building. The activities performed typically has a sequence structure. For example, the person might be performing activities such as bathing, cooking, eating, sleeping {\it etc.} in a sequential order. As in any classification setting, the sequence labeling domain also has complex relationships among inputs with uncertainty in these relationships. For example, if sensors at the microwave oven ($microwave$) and those near the plates in a cupboard ($platesCupboard$) fire within a short time period, it is likely that the person is preparing dinner ($preparingDinner$). Efficient models can be constructed by exploiting these relationships. However, discovering relationships that enhance the discriminative power of classifiers is a hard task, since the space of possible relationships is often exponentially large. Therefore, most of the previous work in structured output space classification and in particular, sequence labeling, either ignore complex input relationships or use heuristics to learn the relationships. In this work, we focus on exploiting complex relationships in an efficient manner in the input as well as the output space to improve sequence labeling models. We begin with  a brief introduction to the task of sequence labeling.

The objective in sequence labeling is to assign a class label to every instance in a sequence of observations (inputs). In general, sequence labeling algorithms learn probabilistic information about transition relationships between neighboring labels (for example, $preparingDinner$ is likely to be followed by $Eating$) along with probabilistic information about the emission/observation relationships between labels and observations (for example, $microwave$ being turned on is related to $preparingDinner$). In a typical non-intrusive activity recognition setting, binary sensor values are recorded at regular time intervals. The joint state of these sensor values at each time instance forms the observation (basic input features). The user activity at a particular time forms the hidden state or label. The history of sensor readings and (manually) annotated activities can be used to train prediction models which could later be used to predict activities based on sensor observations~\cite{kasteren08,naveenActivity}. Hidden Markov Models (HMM)~\cite{hmm}, Conditional Random Fields (CRF)~\cite{CRFLafferty} and Support Vector Machines on Structured Output Spaces (StructSVM)~\cite{AltHofTso06,TsochantaridisSVMStructured} are the most popular approaches to sequence labeling. 

Recent works have shown that learning the relational structure between input features improves the efficiency of sequence labeling models~\cite{McCallum2003,naveenActivity}. For example, in activity recognition settings, one could expect certain combinations of (sensor) readings to be directly indicative of certain activities (for example, $microwave$ and $platesCupboard$ sensors firing within a short time period could be a strong indicator of the activity $preparingDinner$). While HMM, CRF and StructSVM attempt to indirectly capture these relations through their linear combinations, we intend to learn these relations to improve labeling accuracy. However, the space of relational features is exponential in the number of basic inputs, making the discovery of useful features a difficult task. For instance, in creating features from conjunctions of basic inputs at any single sequence position (for example, $microwave$ and $platesCupboard$ firing at the same time step), the feature space is of size {\footnotesize $2^N$} for {\footnotesize $N$} basic inputs. The problem is further exacerbated if we consider complex relational features built from inputs at different relative positions (for example, $platesCupboard$ firing 1 minute after $microwave$ became on). An exhaustive search in this exponentially large feature space is infeasible. Therefore, most systems that learn relational features employ greedy search strategies based on heuristics. In this paper, we explore the possibility of optimally learning and using discriminative relational features (relational features that discern labels) for sequence labeling. That is, discriminative non-linear features, that are indicative of each label, have to be learned. %\textbf{Ajay: consider rewording the above sentence, or drop the sentence and merge the discriminative non-linear features to the previous sentence}. 
For example, $microwave$ and $platesCupboard$ make the activity $preparingDinner$ more likely. This can be viewed as decision rules of the form {\it if condition then decision}, where {\it decision} part is a single variable representing a label to be classified and the {\it condition} part is a composition (possibly conjunction) of a small number of simple boolean statements concerning the values of the individual input variables. This type of a rule, with a single boolean predicate as the decision variable, is generally referred to as a \emph {definite clause}. From a definite clause representation, a label-specific feature can be extracted from the condition part (which can be a conjunction of multiple literals) of the definite clause whose head depicts the class label. Therefore, we use the terms (definite) {\it clause} and (label specific) {\it feature} interchangeably. For simplicity, we restrict our discussion to function-free definite clauses.

To formalize the objective of learning relational features, we categorize relational features based on their complexity (and utility) and show that certain categories of complex features can be constructed from simpler ones. We identify feature categories that are relevant and useful for discriminative prediction models and develop optimal learning approaches for those categories.

An outline of our optimal learning approaches for the useful feature categories is as follows: to begin with, we learn features that are conjunctions of basic inputs at a single sequence position, which is the feature space explored in \cite{naveenActivity} and \cite{CRFMcCallum}. We propose a Hierarchical Kernels-based approach for discovering features from this space. Hierarchical Kernel Learning (HKL) was originally introduced by Bach \cite{bachHKL} for high-dimensional non-linear variable selection, by exploiting the hierarchical structure of the problem. \cite{ganeshRELHKL} leveraged the HKL framework to learn rule ensembles for binary classification tasks. Our approach, referred to as Hierarchical Kernel Learning for Structured Output Spaces (StructHKL)\footnote{\scriptsize This part of our work has appeared at AAAI, 2012~\cite{aaaiNaveen}.}, optimally and efficiently learns discriminative conjunctive features for multi-class structured output\footnote{\scriptsize Structured output where each node in the structure can possibly be labeled as one of multiple class/label} classification problems such as sequence labeling. We build on the StructSVM framework  %~\cite{TsochantaridisSVMStructured,AltHofTso06} 
for sequence prediction problems, in which all possible conjunctions form the input features while the transition features are constructed from all possible transitions between state labels. We use a $\rho$-norm hierarchical regularizer to select a sparse set of emission features.% Since we need to preserve all possible transitions, a conventional 2-norm regularizer is employed for state transition features.
The exponentially large observation feature space is searched using an active set algorithm and the exponentially large set of constraints is handled using a cutting plane algorithm.

As stated before, we further study the possibility of efficiently learning and using discriminative relational features, that capture sequential information among input variables for sequence labeling. (for example, $platesCupboard$ firing 1 minute after $microwave$ is turned on). The applicability of StructHKL in learning complex relational features that are derived from inputs at different relative positions in a sequence, is non-trivial and challenging. Therefore, from our feature categorization, we identify simpler categories that can be {\em AND}ed to yield complex ones, with the goal of formulating efficient, yet effective relational feature learning procedures. Therefore, optimal relational discriminative features can be learned either by (i) enumerating simpler component features and discovering their useful conjunctions using StructHKL or by (ii) developing methods to learn complex relational features directly and optimally.

For the former option above, the space of component features itself could be prohibitively large, and therefore, it may not be feasible to enumerate all such features in a domain. Therefore, we propose to selectively enumerate component features based on certain relevance criteria, such as {\em support} of the feature in the training set.

Once a relevant set of component features are generated, StructHKL can be employed to learn their conjunctions \footnote{\scriptsize This part of our work has appeared at ILP, 2012 \cite{ilp2012paper}.}. Since selective enumeration of component features is not guaranteed to be optimal, this approach is optimal only with respect to the selected components. We propose a relational subsequence kernel-based approach to implicitly learn optimal relational features. Relational subsequence kernels~\cite{subSequenceKernels} compute the similarity between all possible subsequences, which implicitly covers the relational feature space. While this method of modeling does not result in interpretability, relational subsequence kernels do efficiently capture information in entire relational feature space\footnote{\scriptsize This part of our work has appeared in ILP 2013~\cite{ilp2013Paper}.}.

We perform extensive empirical evaluation on publicly available activity recognition datasets and record our observations on settings in which certain approaches are effective. The paper is organized as follows. Section~\ref{sec:related} discusses background work. In Section~\ref{sec:firstorderfeatures}, the complexity based categorization of features is discussed. We discuss our approaches in Section~\ref{sec:featureInduction}. Experimental setup and results are discussed in Section~\ref{sec:experiments} and we conclude the paper in Section~\ref{sec:conclusion}.

\section{Background}\label{sec:related} 

The relevant background work can be broadly categorized into two types: (i) probabilistic and max-margin models for sequence labeling (ii) learning input relationships for building efficient models, which includes the approaches for learning relationships in sequence labeling and an optimal learning approach that has been recently developed for binary classification settings. We discuss each of these in the subsections to follow.

\subsection{Models for Sequence Labeling}

In sequence labeling models, the objective is to learn functions of the form {\footnotesize $\mathcal{F}:\mathcal{X}\rightarrow \mathcal{Y}$} from the training data, where {\footnotesize $\mathcal{X}$} and {\footnotesize $\mathcal{Y}$} are input and output sequence spaces, respectively. Typically, a discriminant function {\footnotesize $F:\mathcal{X} \times \mathcal{Y} \rightarrow \mathbb{R}$} is learned from training data that consists of pairs of input and output sequences. The prediction is performed using the decision function {\footnotesize $ \mathcal{F}(X;\mathbf{f})=\argmax\limits_{Y\in\mathcal{Y}} F(X,Y;\mathbf{f})$}, where {\footnotesize $F(X,Y;\mathbf{f}) = \langle \mathbf{f},\boldsymbol{\psi}(X,Y) \rangle$} represents a score which is a scalar value based on the features {\footnotesize $\boldsymbol{\psi}$} involving the values of input sequence {\footnotesize $X$} and output sequence {\footnotesize $Y$}\footnote{\scriptsize $\boldsymbol{\psi}(X,Y)$ stands for features representing the characteristics of input and output variables, and their relations. For example, a feature can be the state of a sensor at time $t$ or the joint state of two sensors at time $t$.} and parameterized by a parameter vector $\mathbf{f}$. In sequence prediction, features are constructed to 
represent emission (observation) and transition distributions. Given this objective, we can classify sequence labeling 
techniques into probability based and max-margin based, which we discuss in the following paragraphs.

\noindent\textbf{Probability-based methods} : Here the parameters are characterised by probabilities. Hidden Markov Models (HMM)~\cite{hmm} and Conditional Random Fields (CRF)~\cite{CRFLafferty} are traditionally used in sequence prediction problems and have a similar objective as discussed above -- with probabilities and potential weights as parameters, respectively. Their ability to capture the state transition dependencies along with the observation dependencies makes these approaches robust in noisy\footnote{\scriptsize For example, activity recognition data is noisy due to faulty sensors, communication lines and non-uniform activity patterns adopted by subjects to perform tasks.} and sparse data.

In an HMM setup, probability parameters that maximise the joint probability {\footnotesize $p(X,Y)$}, of input training sequence {\footnotesize $X$} and output training sequence {\footnotesize $Y$}, are learned during the training phase. From the independence assumptions in an HMM, one can factorise the joint probability distribution of the sequence of inputs ({\footnotesize $X$}) and labels ({\footnotesize $Y$}) into three factors: the initial state distribution {\footnotesize $p(y^1)$}, the transition distribution {\footnotesize $p(y^t| y^{t-1})$}, and the emission distribution {\footnotesize $p(x^t| y^t)$}~\cite{hmm}. Here {\footnotesize $x^t$} and {\footnotesize $y^t$} represent the input variable and the class variable at time {\footnotesize $t$}, respectively. Therefore, {\footnotesize $p(X,Y) = \prod_{t=1}^T p(y^t| y^{t-1}) p(x^t| y^t)$}, where {\footnotesize $T$} is the length of the sequence and {\footnotesize $p(y^1| y^0)$} is used instead of {\footnotesize $p(y^1)$} to simplify notation. Parameters for the distributions are 
learned by maximising {\footnotesize $p(X,Y)$}.
In contrast, CRF~\cite{CRFLafferty} learns parameters that maximise {\footnotesize $p(Y|X)$}, the conditional probability of a sequence of states {\footnotesize $Y$} given a sequence of inputs {\footnotesize $X$}, where, {\footnotesize $p(Y|X) = \frac{1}{Z(X)}~exp\sum\limits_{t=1}^T~\phi_t(y^t,X)+\phi_{t-1}(y^{t-1},y^t,X)$}, in which {\footnotesize $\phi_t(y^t,X)$} and {\footnotesize $\phi_{t-1}(y^{t-1},y^t,X)$} stands for potential functions and {\footnotesize  $Z(X)$} is the partition function.
These parameters are later used to identify the (hidden) label sequence that best explains a given sequence of inputs (or observations) -- which is usually performed by a dynamic programming algorithm called the Viterbi Algorithm~\cite{viterbi}. 

\noindent\textbf{Max-margin based method}: \cite{TsochantaridisSVMStructured} generalizes the SVM framework to perform classification on structured outputs. This builds on the conventional SVM formulation that assumes the output to be a single variable, which can be a binary label or multi-class. StructSVM generalizes multi-class Support Vector Machine learning to incorporate features constructed from input and output variables and solves classification problems with structured output data. The loss function in sequence labeling has to be chosen such that, the predicted sequence of labels that differ from the actual labels in a few sequence positions should be penalized less than those that differ from the actual labels in a majority of sequence positions. We use average of wrong labelings as loss function in our derivations and experiments\footnote{\scriptsize Fraction of sequence positions wrongly labeled.}. A loss function is represented as {\footnotesize $\Delta:\mathcal{Y}\times \mathcal{Y} \rightarrow \mathbb{R}$}. {\footnotesize $\Delta(Y,\hat{Y})$} is the loss value when the true output is {\footnotesize $Y$} and the prediction is {\footnotesize $\hat{Y}$}. The SVM formulation for structured output spaces can be written as,
 
\vspace{-0.45cm}
{\footnotesize 
\begin{IEEEeqnarray*}{lCl}
 \min_{\mathbf{f},\boldsymbol{\xi}}\; \frac{1}{2}\parallel \mathbf{f}\parallel ^2 \; + \; \frac{C}{m}\sum\limits_{i=1}^m\xi_i,\quad \quad s.t. \;\;\forall i:\; \xi_i \geq 0 \\
\forall i,\;\forall\; Y \in \mathcal{Y}\setminus Y_i :\; \langle \mathbf{f}, \boldsymbol{\psi}_i^{\delta}(Y) \rangle \geq 1 - \frac{\xi_i}{\Delta(Y_i,Y)}.\IEEEyesnumber
\label{svm0}
\end{IEEEeqnarray*}
}
where {\footnotesize $m$} is the number of examples, {\footnotesize $C$} is the regularisation parameter, {\footnotesize $\xi$}'s are the slack variables introduced to allow errors in the training set in a soft margin SVM formulation and  {\footnotesize $X_i \in \mathcal{X}$} and  {\footnotesize $Y_i \in \mathcal{Y}$} represent the {\footnotesize $i^{th}$} input and output sequences, respectively,
\footnote{Subscript $i$ here is to denote $i^{th}$ example sequence and should not be confused with the $i^{th}$ element of a vector}.
{\footnotesize $\langle \mathbf{f}, \boldsymbol{\psi}_i^{\delta}(Y) \rangle$} represents the value {\footnotesize $\langle \mathbf{f}, \boldsymbol{\psi}(X_i,Y_i) \rangle - \langle \mathbf{f}, \boldsymbol{\psi}(X_i,Y) \rangle$}.
In cases where the sequence length is large, the number of constraints in (\ref{svm0}) can be extremely large. To solve this problem, an algorithm based on the cutting plane method is proposed by \cite{TsochantaridisSVMStructured} ({\it c.f.} algorithm 1 in~\cite{TsochantaridisSVMStructured}) to find a polynomially-sized subset of constraints that ensures a solution very close to the optimum.

In this paper, we build on StructSVM to learn relational structure in input space for improving labeling accuracy. The next section discusses some of the existing approaches that examine the learning of relationships to improve sequence labeling.

\subsection{Learning Relationships as Features}

First, we discuss greedy feature induction approaches for sequential data and then discuss an optimal feature induction approach that works for binary classification problems.
McCallum \cite{CRFMcCallum} as well as our prior work~\cite{naveenActivity}, propose feature induction methods that iteratively construct feature conjunctions that increase an objective. These approaches start with an initial set of features, and at each step, consider a set of candidate features (conjunctions or atomic). Features whose inclusion will lead to a maximum increase in the objective are selected. Weights for the new features are trained. The steps are iterated until convergence. However, the approach  does not learn relational features (if ground basic inputs are not provided or not feasible to provide) from multiple relative sequence positions. While \cite{CRFMcCallum} trains a CRF model and uses conditional log-likelihood as the score for greedy induction, we used HMM evaluation on a held out data as the score. This effectively helps to capture non-linear relationships among inputs without representing them in an exponential observation space. However, being greedy, these approaches do not guarantee optimality.

TildeCRF~\cite{tildeCRF} is another approach that learns higher order features for sequences. In this approach, the relational structure and parameters of a CRF for sequence labeling are learned. TildeCRF uses relational regression trees and gradient tree boosting for learning the structure and parameters. However, this approach does not guarantee that the learned structure is optimal.

Jawanpuria {\it et al.}~\cite{ganeshRELHKL} propose Rule Ensemble Learning using Hierarchical Kernels where they make use of the Hierarchical Kernel Learning (HKL) framework introduced in ~\cite{bachHKL} to simultaneously learn sparse rule ensembles and their optimal weights for binary classification problems. We will refer to their approach as RELHKL. They use a hierarchical $\rho$-norm ($\rho \in (1,2]$) regularizer to select a sparse set of features from an exponential space of features. We briefly discuss their approach in the following paragraphs.

The prime objective of Rule Ensemble Learning (REL) is to learn a small set of simple rules and their optimal weights. The set of rules that can be constructed from basic features follow a partial order and can be visualised as a lattice (conjunction lattice when the features are conjunctions of basic features). The set of indices of the nodes (conjunctions) in the lattice are represented by {\footnotesize $\mathcal{V}$}. To learn sparse sets of rules, the regularizer {\footnotesize $\Omega(\mathbf{f})$} is defined as~\cite{ganeshRELHKL},
{\footnotesize $\Omega(\mathbf{f})=\sum_{v\in\mathcal{V}}d_v\parallel \mathbf{f}_{D(v)}\parallel _\rho$}, where {\footnotesize $\mathbf{f}$} is the feature weight vector corresponding to the feature nodes in the lattice, {\footnotesize $d_v\ge0$} is a prior parameter showing the usefulness of the feature conjunctions, {\footnotesize $\mathbf{f}_{D(v)}$} is the vector with elements as {\footnotesize $\parallel f_w\parallel _2 \;\forall w \in D(v)$, $D(v)$} the set of descendant nodes of {\footnotesize $v$} and {\footnotesize $\parallel .\parallel _\rho$} represents the $\rho$-norm. In rule ensemble learning {\footnotesize $d_v$} is defined as {\footnotesize $\beta^{|v|}$}, where {\footnotesize $\beta$} is a constant. Here, the 1-norm over the sub lattices formed by the descendants of nodes helps to select only a few sub lattices and the $\rho$-norm ensures sparsity among the selected sub lattices \cite{ganeshRELHKL}.
Since only a few features are expected to be non-zero at optimality, for computational efficiency, an active set algorithm is employed ({\it c.f.} algorithm 1 of~\cite{ganeshRELHKL}).

The RELHKL approach is specific to the single variable binary classification problem and cannot be trivially applied to problems involving multi class structured output data. In Section \ref{sec:featureInduction}, we present our generalisation of the RELHKL formulation for structured output spaces and discuss how interesting relational features can be learned using our new framework.

\section{Definite Clauses/Features}
\label{sec:firstorderfeatures}

As stated in the introduction, our objective is to learn definite clauses that define rules/features. We start by defining categories of predicates and then discuss the complexity-based classification of features.
Similar to the {\it structural} and {\it property} predicates in 1BC clauses~\cite{1BC}, we define two types of predicates, namely, {\it (inter)  relational} and {\it evidence} predicates. A {\it relational} predicate is a binary predicate that represents the relationship between {\it types} or between a {\it type} and its parts. A {\it type} is an entity or object that has a meaning and described by itself or its attributes. An {\it evidence} predicate is an assertion of a situation or a property of a {\it type} or part of it\footnote{\scriptsize Evidence predicate here is not the same as that used in Markov Logic Networks, where it means observed variable. A property is some conclusive information about a type.}. For example, \texttt{microwave(T)} states that the microwave was switched on at time \texttt{T}. We use the following set of definite clause examples for illustration of the concepts we discuss in the the rest of this section.

{
\begin{enumerate}
\small{
\item {\texttt{{preparingDinner(T)  :- microwave(T)}}}
 \item \texttt{{preparingDinner(T)  :- microwave(T), platesCupboard(T)}}
\item {\texttt{{preparingDinner(T1):- prevRelPosWindowNear(T1,T2), platesCupboard(T2)}}}
 \item \texttt{{preparingDinner(T1):- prevRelPosWindowNear(T1,T2)}}
 \item {\texttt{{preparingLunch(T1) :- prevRelPosWindowNear(T1,T2), platesCupboard(T2), microwave(T2)}}}
 \item {\texttt{{preparingLunch(T1) :- prevRelPosWindowNear(T1,T2), platesCupboard(T2), prevRelPosWindowNear(T1,T3), microwave(T3), greater(T2,T3) }}}
\item \texttt{{preparingDinner(T1) :- microwave(T1), prevRelPosWindowNear(T1,T2), platesCupboard(T2)}}
}
\end{enumerate}
}

Based on complexity, we categorize definite features as \emph{simple conjunctions} ($\mathcal{SC}$), \emph{absolute features} ($\mathcal{AF}$), \emph{primary features} ($\mathcal{PF}$), \emph{composite features} ($\mathcal{CF}$) and \emph{definite features} ($\mathcal{DF}$).\\

\noindent{\bf Simple Conjuncts ($\mathcal{SC}$):}

$\mathcal{SC}$s are simple conjunctions of basic features (including unary conjunctions) observed at a single sequence position. In other words, $\mathcal{SC}$s are conjunctions of only evidence predicates. Clauses 1 and 2 above are simple conjunctions.\\

\noindent{\bf Absolute Features ($\mathcal{AF}$):}

In an absolute feature, new local variables can be introduced only in a {\it relational} predicate, where, a local variable is a variable that does not appear in the head predicate. Unlike the 1BC~\cite{1BC} clauses, any number of new local variables can be introduced in a {\it relational} predicate. Any number of {\it relational} and {\it evidence} predicates can be conjoined to form an $\mathcal{AF}$, such that, the resultant $\mathcal{AF}$ is minimal and the local variables introduced in the {\it relational} predicates are (transitively) consumed by some {\it evidence} predicates. In this setting, a minimal clause is one which cannot be constructed from smaller clauses that share no common local variables. Thus, clauses 1, 3, 5 and 6 above are $\mathcal{AF}$s, whereas, clauses 2 (not minimal), 7 (not minimal) and 4 (since variable \texttt{{T2}} is not consumed) are not.\\

\noindent{\bf Primary Features ($\mathcal{PF}$):}

Primary features are absolute features, in which a new local variable introduced is consumed only once.\footnote{\scriptsize This is similar to elementary features in~\cite{1BC} except that elementary features allow only one new local variable in a {\it structural} predicate. Also $\mathcal{PF}$s are different from simple clauses~\cite{limeSimpleClause}, as a simple clause allows a local variable to be unconsumed.}. Only clause 1 and 3 are $\mathcal{PF}$s.\\

\noindent{\bf Composite Features ($\mathcal{CF}$):}

Composite Features are features formed by the conjunction of one or more $\mathcal{AF}$s without unification of body literals. Only the head predicates are unified. 
Clauses 1 , 2, 3 , 5 , 6 and 7 are $\mathcal{CF}$s, whereas, clause 4 is not.\\

\noindent{\bf Definite Features ($\mathcal{DF}$):}

Definite features are features with none of the above restrictions. Therefore, all the given examples are $\mathcal{DF}$s.\\

We now state some relationships between these feature categories. \\%The proofs are included in the appendix.\\

\noindent {\bf Relationships among Feature Categories}
\begin{enumerate}
 \item ${\mathcal PF} \subset {\mathcal AF}$.
  \item ${\mathcal AF} \subset {\mathcal CF}$.
  \item ${\mathcal SC} \subset {\mathcal CF}$.
  \item ${\mathcal CF} \subset {\mathcal DF}$.
  \item Every ${\mathcal AF}$ can be constructed from ${\mathcal PF}$s using unifications.
  \item  Every ${\mathcal CF}$ can be constructed from ${\mathcal AF}$s by conjunctions.
  \item ${\mathcal CF}$s are ${\mathcal DF}$s with local variable reuse restriction. 
\end{enumerate}

Having presented some relevant choices for the feature subspace, we next discuss our approaches for inducing relational features for sequence labeling.

\section{Learning Discriminative Features for Sequence Labeling}
\label{sec:featureInduction}

We recall that our overall objective is to exploit complex relationships among input variables in sequence labeling problems to improve the accuracy of classification, while striving towards efficient discovery of the features. Now, we formalize our intuition and present our proposed approaches in detail.
As stated in Section \ref{sec:related}, the training objective in sequence labeling can be posed as learning feature weights that make  the score {\footnotesize $F(X,Y;\mathbf{f}) = \langle \mathbf{f},\boldsymbol{\psi}(X,Y) \rangle$} of the true output sequence {\footnotesize $Y$} greater than any other possible output sequence, given an input sequence {\footnotesize $X$}.

In Section~\ref{sec:firstorderfeatures}, we have categorized relational features into  simple conjunctions ($\mathcal{SC}$), absolute features ($\mathcal{AF}$), primary features ($\mathcal{PF}$), composite features ($\mathcal{CF}$) and definite features ($\mathcal{DF}$). Of these, $\mathcal{SC}$s, that are features derived from inputs at a single sequence position ($microwave(T) \wedge platesCupboard(T)$), and $\mathcal{CF}$s, that are relational features derived from multiple sequence positions ($microwave(T1) \wedge prevRelPosWindowNear(T1,T2) \wedge platesCupboard(T2)$), are particularly interesting in our problem. 

In this section, we formally discuss our proposed feature learning approaches for these categories. We start with learning features from the space of simple conjunctions ($\mathcal{SC}$). The objective is to learn $\mathcal{SC}$s relevant to each label\footnote{\scriptsize Features whose truth values make the label less/more likely}.  For this purpose, we propose the Hierarchical Kernels-based approach for optimally learning $\mathcal{SC}$s, which we refer to as Hierarchical Kernel Learning for Structured Output Spaces (StructHKL). This is followed by our investigations into the class of complex relational features ($\mathcal{AF}$s and $\mathcal{CF}$s) that capture sequential information among input variables to build efficient sequence labeling models.

\subsection{Hierarchical Kernel Learning for Structured Output Spaces}
\label{sec:structrelhkl}

The space of simple conjunctions ($\mathcal{SC}$) forms a lattice (for the subset relation). The top node is the empty node, nodes at level 1 are all possible basic inputs at any sequence position, nodes at level 2 are composed of two basic inputs and the bottom node is the conjunction of all basic inputs. Therefore, each node is a conjunction of zero or more basic features. In this subsection, we derive our approach in a general setting of features (not necessarily conjunctions) that follow a partial order and then discuss the possibility of learning $\mathcal{SC}$s using this approach. The size of such an ordering is {\footnotesize $2^N$} for {\footnotesize $N$} basic inputs. Our approach leverages the sparsity-inducing regularizer of RELHKL~\cite{ganeshRELHKL} in a StructSVM framework, to discover optimal discriminative features from the partial order for solving sequence labeling problems.

We build on the StructSVM model for sequence prediction problems given in {\it equation} (\ref{svm0}). We modify the feature vector to include all possible label-specific features (that are compositions of basic features), while preserving all possible transitions.  First, we give a brief introduction to the notations we use.

Let the input (observation) at {\footnotesize $p^{th}$} sequence step of the {\footnotesize $i^{th}$} example be {\footnotesize $\mathbf{x}_i^p$}, where {\footnotesize $\mathbf{x}_i^p$} is a vector of binary values. Each element of the vector represents the value of an input at the position {\footnotesize $p$}. For instance, in activity recognition, these binary values represent the values of sensors fixed at locations such as $microwave$, $platesCupboard$ at the {\footnotesize $p^{th}$} time step. Similarly, output (label) at the {\footnotesize $p^{th}$} time step of the {\footnotesize $i^{th}$} example is represented by {\footnotesize $y_i^p$} which can take any of the {\footnotesize $n$} values ($preparingDinner$, $sleeping$, {\it etc.}). We use {\footnotesize $l_i$} to represent the length of sequence {\footnotesize $i$}. Basic or derived features and feature weights are represented by {\footnotesize $\boldsymbol{\psi}$} and {\footnotesize $\mathbf{f}$}, respectively. Elements of {\footnotesize $\boldsymbol{\psi}$} correspond to emission (basic input/
observation) and transition features.  We represent the emission and transition parts of the vector {\footnotesize $\boldsymbol{\psi}$} as {\footnotesize $\boldsymbol{\psi_E}$} and {\footnotesize $\boldsymbol{\psi_T}$}, respectively. We assume that both {\footnotesize $\boldsymbol{\psi_E}$} and {\footnotesize $\boldsymbol{\psi_T}$} are vectors of dimension equal to the dimension of {\footnotesize $\boldsymbol{\psi}$} with zero values for all elements not in their context. That is, {\footnotesize $\boldsymbol{\psi_E}$} has a dimension of {\footnotesize $\boldsymbol{\psi}$}, but has zero values corresponding to the transition elements. Likewise, we split the feature weight vector {\footnotesize $\mathbf{f}$} into {\footnotesize $\mathbf{f_E}$} and {\footnotesize $\mathbf{f_T}$}. Similarly, {\footnotesize $\mathcal{V}$}, the indices of the elements of {\footnotesize $\boldsymbol{\psi}$}, is split into {\footnotesize $\mathcal{V}_\mathbf{E}$} and {\footnotesize $\mathcal{V}_\mathbf{T}$}. As followed in~\cite{ganeshRELHKL}, {\footnotesize $D(v)$} 
and {\footnotesize $A(v)$} represent the set of descendants and ancestors of the node {\footnotesize $v$} in the lattice, respectively. Both {\footnotesize $D(v)$} and {\footnotesize $A(v)$} include node {\footnotesize $v$}. The hull and the sources of any subset of nodes {\footnotesize $\mathcal{W}\subset\mathcal{V}$} are defined as {\footnotesize $hull(\mathcal{W})=\bigcup_{w\in\mathcal{W}} A(w)$} and {\footnotesize $sources(\mathcal{W})=\{w\in\mathcal{W}|A(w)\bigcap\mathcal{W}=\{w\}\}$}, respectively. The size of set {\footnotesize $\mathcal{W}$} is denoted by {\footnotesize $|\mathcal{W}|$. $\mathbf{f}_{\mathcal{W}}$} is the vector with elements as {\footnotesize $f_v, v \in \mathcal{W}$}.   Let the complement of {\footnotesize $\mathcal{W}$} denoted by {\footnotesize $\mathcal{W}^c$} be the set of all features belonging to the same label that are not in {\footnotesize $\mathcal{W}$}. Our proposed approaches are to discover observation features ({\footnotesize $\boldsymbol{\psi_E}$}) capturing complex input relationships. For the 
sake of visualisation, we assume there is a partial order for each label. Therefore, elements of the vector {\footnotesize $\boldsymbol{\psi_E}$} correspond to the nodes in the partial ordering of features for each label.

To use the hierarchical $\rho$-norm regularizer on the feature weights corresponding to the emission nodes, we separate the regularizer term into those corresponding to emission and transition features. Since all the state transitions are to be preserved, a conventional 2-norm regularizer is used for transition features. The new SVM formulation is, 

\vspace{-0.45cm}
{\footnotesize
\begin{IEEEeqnarray*}{lCl}
 \min\limits_{\mathbf{f},\boldsymbol{\xi}}\frac{1}{2}\Omega_E(\mathbf{f_E})^2 \; + \; \frac{1}{2}\Omega_T(\mathbf{f_T})^2 \; +\;  \frac{C}{m}\sum\limits_{i=1}^m\xi_i,\\
\forall i,\forall Y \in \mathcal{Y}\setminus Y_i :\; \langle \mathbf{f}, \boldsymbol{\psi}_i^{\delta}(Y) \rangle \geq 1 - \frac{\xi_i}{\Delta(Y_i,Y)};~~~
\forall i:\; \xi_i \geq 0 \IEEEyesnumber
\label{svm1}
\end{IEEEeqnarray*}
}
where {\footnotesize $\Omega_E(\mathbf{f_E})$} is defined as {\footnotesize $\sum\limits_{v\in \mathcal{V}_\mathbf{E}} d_v \parallel \mathbf{f}_{\mathbf{E}D(v)}\parallel _\rho ,\; \; \rho \in (1,2]$} \normalsize  and  {\footnotesize$\Omega_T(\mathbf{f_T})$} is the 2-norm regularizer {\footnotesize $\big(\sum\limits_j f_{Tj}^2\big)^{\frac{1}{2}}$}\normalsize

The 1-norm in {\footnotesize$\Omega_E(\mathbf{f_E})$} ensures sparsity among many of the {\footnotesize $\parallel\mathbf{f}_{\mathbf{E}D(v)}\parallel _\rho$}. The $\rho$-norm ensures sparsity among nodes that are selected by the 1-norm \cite{ganeshRELHKL}. As the transition feature space is not exponential, sparsity is not desired, and therefore, a 2-norm regularizer is sufficient for transition. This SVM setup is hard to solve because of the following issues: (i) The regularizer, {\footnotesize $\Omega_E(\mathbf{f_E})$}, consists of $\rho$-norm over descendants of each lattice node, which makes it exponentially expensive. (ii) The number of constraints in the formulation is exponential. The rest of the section discusses how to solve the problem efficiently.

By solving (\ref{svm1}), we expect most of the emission feature weights to be zero. %As illustrated by Jawanpuria {\it et al.}, 
Therefore, at optimality, the solution with selected features and the solution with all the features, are same. Hence, for computational efficiency, an active set algorithm can be employed. Since the constraint set in (\ref{svm1}) is exponential, in each iteration of the active set algorithm,  a cutting plane algorithm has to be used. The algorithm finds a polynomial sized subset of constraints so that the corresponding solution satisfies all the constraints with an error not more than {\footnotesize $\epsilon$}. Now, we modify (\ref{svm1}) to consider only the active set of features {\footnotesize $\mathcal{W}$}.

\vspace{-0.45cm}
{\footnotesize
\begin{IEEEeqnarray*}{lCl}
\footnotesize
 \min\limits_{\mathbf{f},\boldsymbol{\xi}}\frac{1}{2}\left(\sum\limits_{v\in \mathcal{W}} d_v \parallel \mathbf{f}_{\mathbf{E}D(v)\bigcap\mathcal{W}}\parallel _\rho \right)^2 \; + \; \frac{1}{2} \parallel \mathbf{f_T}\parallel _2^2 \; +\;  \frac{C}{m}\sum\limits_{i=1}^m\xi_i,\\
\forall i,\forall Y \in \mathcal{Y}\setminus Y_i :~ -\left(\sum\limits_{v\in\mathcal{W}}\langle f_{Ev}, \psi_{Evi}^{\delta}(Y) \rangle +  \sum\limits_{v\in\mathcal{V}_\mathbf{T}}\langle f_{Tv},\psi_{Tvi}^{\delta}(Y) \rangle  +  \frac{\xi_i}{\Delta(Y_i,Y)}  -  1 \right) \leq 0\\
\forall i:\; -\xi_i \leq 0 \IEEEyesnumber
\label{svm2}
\end{IEEEeqnarray*}
}

\vspace{-0.45cm}
\noindent where {\footnotesize $\rho \in$ (1,2]}. The active set algorithm can be terminated when the solution to the small problem (reduced solution) is the same as the solution to the original problem, otherwise, the active set has to be updated. We follow an approach similar to that in~\cite{ganeshRELHKL} to derive a sufficiency condition to check optimality. From the duality gap, we derive the sufficiency condition for the reduced solution with {\footnotesize $\mathcal{W}$} to have a duality gap less than {$\epsilon$} as,

\vspace{-0.45cm}
{\footnotesize
\begin{IEEEeqnarray*}{rCl}
\footnotesize
\underset{u\in sources(\mathcal{W}^c)}{\max}\sum_{i,Y\neq Y_i}\sum_{j,Y'\neq Y_j} \alpha_{\mathcal{W}iY}^\top 
\sum_{p=1}^{l_i}\sum_{q=1}^{l_j} 2 \Big( \prod_{k\in u}\frac{\psi_{Ek}(\mathbf{x}_i^p)\psi_{Ek}(\mathbf{x}_j^q)}{\beta^2}\Big)\\\Big( \prod_{k\not\in u}\big( 1 + \frac{\psi_{Ek}(\mathbf{x}_i^p)\psi_{Ek}(\mathbf{x}_j^q)}{(1+\beta)^2}\big) \Big)    \alpha_{\mathcal{W}jY'}\\ \boldsymbol{\leq} \Omega_E(\mathbf{f}_{\mathbf{E}\mathcal{W}})^2 + \Omega_T(\mathbf{f}_{\mathbf{T}\mathcal{W}})^2 +2(\epsilon - e_{\mathcal{W}})
\IEEEyesnumber
\label{sufficiency}
\end{IEEEeqnarray*}
}

\vspace{-0.45cm}
\noindent where
 {\footnotesize $e_{\mathcal{W}}=\Omega_E(\mathbf{f}_{\mathbf{E}\mathcal{W}})^2 + \Omega_T(\mathbf{f}_{\mathbf{T}\mathcal{W}})^2 + \frac{C}{m}\sum\limits_i \xi_i + \frac{1}{2}\alpha_\mathcal{W}^{\top}\boldsymbol{\kappa}_\mathbf{T}\alpha_\mathcal{W}-\sum\limits_{i,Y\neq Y_i}\alpha_{_\mathcal{W}iY}$}.

If the solution in any iteration of the active set satisfies the above condition, the algorithm terminates; else the active set is updated by adding the nodes in $sources(\mathcal{W}^c)$ that violate this condition.
We now derive the dual of (\ref{svm1}) as,

\vspace{-0.45cm}
{\footnotesize 
\begin{IEEEeqnarray*}{rCl}
\footnotesize
\min_{\boldsymbol{\eta}\in\Delta_{|\mathcal{V}|,1}} g(\boldsymbol{\eta})
\IEEEyesnumber
\label{finalDualOuter}
\end{IEEEeqnarray*}
}
where {\footnotesize $g(\boldsymbol{\eta})$} is defined as,

\vspace{-0.45cm}
{\footnotesize 
\begin{IEEEeqnarray*}{rCl}
\footnotesize
 \max_{\boldsymbol{\alpha}\in \tau(\mathcal{Y},C)} \sum_{i,Y\neq Y_i}\alpha_{iY} - \frac{1}{2}\boldsymbol{\alpha}^\top \boldsymbol{\kappa}_{\mathbf{T}}\boldsymbol{\alpha} - \frac{1}{2}\Big(\sum_{w \in \mathcal{V}} \zeta_w(\boldsymbol{\eta})(\boldsymbol{\alpha}^\top \boldsymbol{\kappa}_{\mathbf{E}w}\boldsymbol{\alpha})^{\bar{\rho}}\Big)^{\frac{1}{\bar{\rho}}},\\
\IEEEyesnumber
\label{finalDualInner}
\end{IEEEeqnarray*}
}

 \vspace{-0.45cm}
\noindent and {\footnotesize $\zeta_w(\boldsymbol{\eta})=\big(\sum\limits_{v\in A(w)}d_v^\rho \eta_v^{1-\rho}\big)^{\frac{1}{1-\rho}}$} \; and \; {\footnotesize $\bar{\rho}=\frac{\rho}{2(\rho - 1)}$}.

The solution to the dual problem in (\ref{finalDualOuter}) with {\footnotesize $\mathcal{V}$} restricted to the active set {\footnotesize $\mathcal{W}$} yields the solution to the restricted primal problem in (\ref{svm2}). The active set algorithm begins with top nodes in the lattice, solves the dual problem (parameters are updated by solving equation (\ref{finalDualOuter})), checks the sufficiency condition of sources of the complement set of the active set, and adds the nodes that violate the sufficiency condition. The process is continued till no node violates the sufficiency condition. To solve \ref{finalDualOuter}, the sub-gradient is computed as follows.

Let {\footnotesize $\bar{\boldsymbol{\alpha}}$} be the optimal solution to (\ref{finalDualInner}) with some {\footnotesize $\boldsymbol{\eta}$}, then the {\footnotesize $i^{th}$} sub-gradient is computed as

\vspace{-0.45cm}
{\footnotesize 
\begin{IEEEeqnarray*}{rCl}
(\bigtriangledown g(\boldsymbol{\eta}))_i = -~\frac{d_i^{\rho}\eta_i^{-\rho}}{2\bar{\rho}}\Big(\underset{w\in \mathcal{V}_\mathbf{E}}{\sum}\zeta_w(\boldsymbol{\eta})(\bar{\boldsymbol{\alpha}}^\top \boldsymbol{\kappa}_{\mathbf{E}w}\bar{\boldsymbol{\alpha}})^{\bar{\rho}}\Big)^{\frac{1}{\bar{\rho}}-1}
\Big(\underset{w\in D(i)}{\sum}\zeta_w(\boldsymbol{\eta})^{\rho}(\bar{\boldsymbol{\alpha}}^\top \boldsymbol{\kappa}_{\mathbf{E}w}\bar{\boldsymbol{\alpha}})^{\bar{\rho}}\Big)
\IEEEyesnumber
\label{subgradient}
\end{IEEEeqnarray*}
}
To compute the gradient, {\footnotesize $\bar{\boldsymbol{\alpha}}$} is to be obtained by solving (\ref{finalDualInner}). A cutting plane method can be employed for solving (\ref{finalDualInner}). The cutting plane algorithm starts with no constraints for (\ref{svm2}) and in each step, adds a constraint that most violates the margin. The dual problem (\ref{finalDualInner}) is then solved and the process is continued. The process stops when there are no more margin violations. 

In general, StructHKL can be used in structured output classification problems to learn from complex feature spaces that can be ordered as  a lattice and where the summation of descendant kernels can be computed in time polynomial in the number of basic inputs~\cite{bachHKL}. We briefly discuss the possibility of learning $\mathcal{SC}$s using StructHKL.

A simple conjunction ($\mathcal{SC}$) is defined as a conjunction of basic boolean inputs at any single sequence position. The space of conjunctions is a lattice, with an empty node as the top node and the conjunction of all basic inputs as the bottom node. Unlike in~\cite{ganeshRELHKL}, our setting has a conjunction lattice for each label. With an artificial top and bottom node, the whole ordering can still be visualised as a lattice. In the case of $\mathcal{SC}$s, the kernel {\footnotesize $\kappa_v(x_i^p,x_j^q)$} at node {\footnotesize $v$} for sequence positions {\footnotesize $p$} and {\footnotesize $q$} (of examples {\footnotesize $i$} and {\footnotesize $j$}, respectively)  is the kernel induced by the {\footnotesize $v^{th}$} conjunction in the partial order evaluated at the {\footnotesize $p^{th}$} and {\footnotesize $q^{th}$} positions of examples {\footnotesize $i$} and {\footnotesize $j$}, respectively. Therefore, {\footnotesize $\kappa_v(x_i^p,x_j^q)$} is the product of {\footnotesize $v^{th}$} conjunction evaluated at ({\footnotesize $i,p$}) and ({\footnotesize $j,q$}), which is equal to the product of the basic boolean inputs in {\footnotesize $v$} evaluated at ({\footnotesize $i,p$}) and ({\footnotesize $j,q$}). As in~\cite{ganeshRELHKL}, for a sub-space formed by the descendants of a node in the ordering, this sum of products can be written as a product of sums. For instance, {\footnotesize $\sum\limits_{v\in\mathcal{V}} \kappa_v (x_i^p,x_j^q)=\prod\limits_{k\in B}(1+\psi_k(x_i^p)\psi_k(x_j^q))$}, where {\footnotesize $B$} is the set of nodes at level 1 (basic inputs). Therefore, StructHKL can be employed to discover $\mathcal{SC}$s efficiently. 

\subsection{Learning Complex Relational Features for Sequence Labeling}
\label{subsubsec:complexRelationalFeatures}

Our objective is to learn complex relational features, that are derived from inputs at different relative positions. TildeCRF~\cite{tildeCRF} is an existing approach that explores such a feature space, using Inductive Logic Programming (ILP) techniques. However, the approach pursued in it is greedy. We investigate the possibility of leveraging optimal feature learning approaches to explore such a feature space.

Although the StructHKL algorithm optimally solves the objective of learning the most discriminative  $\mathcal{SC}$s for sequence labeling, its applicability in learning complex relational features, that are derived from inputs at different relative positions, is non-trivial and challenging. Therefore, we progressively identify classes of features from the categories discussed in Section \ref{sec:firstorderfeatures}, that can be learned, and analyze the possibility of leveraging StructHKL to optimize learning steps to the extent possible.  We note that the class of composite features ($\mathcal{CF}$) is a powerful set of features capable of representing complex relational information. In Section \ref{sec:firstorderfeatures}, we state that composite features ($\mathcal CF$) can be constructed from absolute features ($\mathcal AF$) with unary or multiple conjunctions without unifications and $\mathcal AF$s can be constructed from primary features ($\mathcal PF$) with unifications. Therefore, the space of $\mathcal CF$s can be defined as a partial order over $\mathcal PF$s with unifications and conjunctions.  However, since $\mathcal{CF}$s (and $\mathcal{AF}$s) share local variables across predicates in the condition part of the clause and the refinement of such a clause is performed by operators such as unification and anti-unification, StructHKL cannot be applied to learn $\mathcal{CF}$s (and $\mathcal{AF}$s). It is easy to observe that, if $\mathcal{AF}$s can be constructed efficiently by other approaches, StructHKL can be employed to efficiently construct $\mathcal{CF}$s from $\mathcal{AF}$s. We identify two possibilities to meet this objective, namely (i) enumerating $\mathcal{AF}$s and discovering their useful compositions ($\mathcal{CF}$) using StructHKL or (ii) developing methods to learn optimal $\mathcal{AF}$s (or $\mathcal{CF}$s directly). We now investigate the former choice of constructing  $\mathcal{CF}$s from a set of enumerated $\mathcal{AF}$s.

\subsubsection{Constructing Composite Features from Enumerated Absolute Features }
\label{subsubsec:CFfromAF}

As $\mathcal{CF}$s are conjunctions of $\mathcal{AF}$s, we can possibly enumerate all $\mathcal{AF}$s and construct an ordering of their conjunctions. StructHKL can be employed to discover $\mathcal{CF}$s from this new ordering.

The space of $\mathcal{AF}$s is prohibitively large, and therefore, it is not feasible to enumerate all $\mathcal{AF}$s in any reasonable domain.   We, therefore, propose to selectively enumerate $\mathcal{AF}$s based on certain relevance criteria such as support of the $\mathcal{AF}$ in the training set. This can be viewed as projecting the space of $\mathcal{CF}$s into the space of $\mathcal{SC}$s\footnote{\scriptsize If we consider each $\mathcal{AF}$ as atomic entity, the conjunctions (which are $\mathcal{CF}$s) can be viewed as $\mathcal{SC}$s formed from these atomic entities} and leveraging StructHKL. With suitable language restrictions, $\mathcal AF$s can be generated using ILP methods. We use a relational pattern miner called \emph{Warmr}~\cite{warmr-paper-1,warmr-paper-2} to generate $\mathcal{AF}$s. Warmr uses a modified version of the Apriori algorithm to find frequent patterns ($\mathcal{AF}$s) which have support exceeding a minimum threshold specified by the user. Once a set of relevant $\mathcal{AF}$s are enumerated, StructHKL can be employed to learn useful compositions of $\mathcal{AF}$s and their parameters to obtain the final model. We now discuss an alternative approach of leveraging relational kernels to implicitly learn relational features.

\subsubsection{Leveraging Complex Relational Kernels for Sequence Labeling}
\label{subsubsec:complexRelationalKernels}

We have defined $\mathcal{CF}$s as features that are derived from a subset of basic attributes at the current position as well as its relative positions (for example, $microwave(T1) \wedge prevRelPosWindowNear(T1,T2) \wedge platesCupboard(T2)$). Since it is not feasible to discover optimal $\mathcal{CF}$s using StructHKL, in the sequence labeling model, we leverage a relational kernel that computes the similarity between instances in an implicit feature space of $\mathcal{CF}$s. To this end, we employ the relational subsequence kernel~\cite{subSequenceKernels} at each sequence position (over a time window of observations around the pivot position) for the classification model. 
Subsequence kernels have been used to extract relations between entities in natural language text \cite{subSequenceKernels}, where the relations are between protein names in biomedical texts. The features are (possibly non-contiguous) sequences of  word and word classes anchored by the protein names at their ends. They extend string kernels \cite{stringKernels} for this task. We briefly discuss relational subsequence kernels in the following paragraph.

Suppose we consider an input {\footnotesize $\mathbf{x}_i^p$} at position {\footnotesize $p$}, for example {\footnotesize $i$}. Let the previous {\footnotesize $k$} positions relative to {\footnotesize $p$} have inputs {\footnotesize $\mathbf{x}_i^{p-1}, \dots \mathbf{x}_i^{p-k}$} and next {\footnotesize $l$} positions relative to {\footnotesize $p$} have inputs {\footnotesize $\mathbf{x}_i^{p+1}, \dots \mathbf{x}_i^{p+l}$}. Let there be {\footnotesize $N$} basic features at a sequence position {\footnotesize $t$} denoted by {\footnotesize $x^{1^t} \dots x^{N^t}$}\footnote{\scriptsize Ignoring the example number $i$ for simplicity}. Essentially our sequence for the particular time-step pivoted at {\footnotesize $p$} denoted by {\footnotesize $Q^p$} is as follows :

\vspace{-0.4cm}
{\footnotesize
\[
 Q^p = \{x^{1^{p-k}}, \dots x^{N^{p-k}} \}, \dots, \{ x^{1^{p-1}}, \dots x^{N^{p-1}} \}, \{ x^{1^p}, \dots x^{N^p} \}, \{ x^{1^{p+1}}, \dots x^{N^{p+1}} \}
\]
}
\vspace{-0.4cm}
{\footnotesize
\[
 {\scriptsize{.}} \qquad \qquad \qquad \qquad \qquad \qquad \qquad \qquad \qquad \qquad \qquad \qquad \qquad \qquad \dots \{ x^{1^{p+l}}, \dots x^{N^{p+l}} \}
\]
}

\noindent Given two sub-sequences {\footnotesize $Q^p$} and {\footnotesize $Q^q$}, we define the relational subsequence kernel {\footnotesize $SSK(Q^p, Q^q) $} as elaborated in \cite{subSequenceKernels}. This kernel is equivalent to the inner product of features that enumerate all possible subsequences in {\footnotesize $Q^p$} and {\footnotesize $Q^q$}. We now show that the feature space of $\mathcal{CF}$ is indeed captured by our relational subsequence kernel. \\

\noindent {\bf Claim:}
Relational subsequence kernels implicitly covers the entire feature space defined by Composite Features ($\mathcal{CF}$) given a constant context window.\\ 
\noindent\textbf{Proof:}\emph{
By definition, the relational subsequence kernel  {\footnotesize $SSK(Q^p, Q^q)$} is equivalent to the inner product of features that enumerate all possible subsequences in {\footnotesize $Q^p$} and {\footnotesize $Q^q$}. Therefore, the kernels implicitly capture all possible common subsequences between {\footnotesize $Q^p$} and {\footnotesize $Q^q$}. $\mathcal{CF}$s are conjunctions of $\mathcal{AF}$s, that are features derived from inputs at the current sequence position and inputs at positions before and after the current position. Since we are considering all the subsequences in the given context (time) window in the relational kernel, we implicitly cover the space of $\mathcal{CF}$s.
}

We now define the kernel for StructSVM framework below, which represents the kernel resulting from the difference in values for the original and the candidate sequences. This stands for the inner product, {\footnotesize $\langle \boldsymbol{\psi_{i}}^\delta(Y),\boldsymbol{\psi_{j}}^\delta(Y^{'})\rangle$} with {\footnotesize $\psi_{j}^\delta(Y)$} defined as {\footnotesize $\psi(X_i,Y_i)-\psi(X_i,Y)$}. The kernel is,

\vspace{-0.45cm}
{\footnotesize 
\begin{IEEEeqnarray*}{lCl}
 \kappa\big((X_i,Y_i,Y),(X_j,Y_j,Y^{'})\big)=\kappa_T(Y_i,Y,Y_j,Y^{'})+\kappa_E\big((X_i,Y_i,Y),(X_j,Y_j,Y^{'})\big)
\label{deltaKernel}
\IEEEyesnumber
\end{IEEEeqnarray*}
}

\vspace{-0.45cm}
where $\kappa_T$ is the transition part of the kernel defined as,

\vspace{-0.45cm}
{\footnotesize 
\begin{IEEEeqnarray*}{lCl}
 \kappa_T\big(Y_i,Y,Y_j,Y^{'})=\kappa_T(Y_i,Y_j)+\kappa_T(Y,Y^{'})-\kappa_T(Y_i,Y^{'})-\kappa_T(Y_j,Y),
\label{deltaTransitionKernel}
\IEEEyesnumber
\end{IEEEeqnarray*}
\vspace{-0.45cm}
\begin{IEEEeqnarray*}{rCl}
\kappa_T(Y_i,Y_j) = \sum\limits_{p=1}^{l_i-1}\sum\limits_{q=1}^{l_j-1} \Lambda(y_i^p,y_j^q)\Lambda(y_i^{p+1},y_j^{q+1})
~~~ = \sum\limits_{p=2}^{l_i}\sum\limits_{q=2}^{l_j} \Lambda(y_i^{p-1},y_j^{q-1})\Lambda(y_i^p,y_j^q),
\label{baseTransitionKernel}
\IEEEyesnumber
\end{IEEEeqnarray*}
}
\noindent {\footnotesize $\Lambda(y_i^p,y_j^q) = 1$} if {\footnotesize $y_i^p=y_j^q$}; {\footnotesize 0} otherwise.
and $\kappa_E$ is the emission part defined as, 
{\scriptsize 
\begin{IEEEeqnarray*}{lCl}
\kappa_E\big((X_i,Y_i,Y),(X_j,Y_j,Y^{'})\big) = \sum\limits_{p=1}^{l_i}\sum\limits_{q=1}^{l_j} \kappa_E(x_i^p,x_j^q)\Big(\Lambda(y_i^p,y_j^q)+\Lambda(y^p,y^{'q})-\Lambda(y_i^p,y^{'q})-\Lambda(y^p,y_j^q)\Big)~~~~~~~~~~~~
\label{deltaEmissionKernel}
\IEEEyesnumber
\end{IEEEeqnarray*}
} 

In our setting of relational subsequence kernels for StructSVM, the kernel {\footnotesize $\kappa_E(x_i^p,x_j^q)$} is the relational subsequence kernel, where we consider a finite window of sequence positions before and after {\footnotesize $p$} and {\footnotesize $q$}, with {\footnotesize $p$} and {\footnotesize $q$} as pivots.

While this method of modelling does not result in interpretability, relational subsequence kernels do efficiently capture the relational sequential information on the inputs. We now discuss our experiments and results.

\section{Experiments}
\label{sec:experiments}

All of our implementation has been in Java. We ran our experiments on a 24-core (2.66 GHz) 64 bit AMD machine with 64 GB RAM and running Ubuntu 11.04. We use three publicly available activity recognition datasets for our experiments. The first is the data provided by Kasteren {\it et al.} \cite{kasteren08}.  The dataset was extracted from a household fitted with 14 binary sensors. Eight activities have been annotated for four weeks. Activity labels are daily house-hold activities like $sleeping$, $usingToilet$, $preparingDinner$, $preparingBreakfast$, $leavingOut$, and others. A data instance was recorded for a time interval of 60 seconds and there are 40006 such data instances. In our experiments, we considered subsequences of length 1000 as examples\footnote{\scriptsize We intuitively chose example size to be 1000, since we wanted our examples to contain enough sequence information and at the same time small enough to conveniently run experiments.}, and there are 40 such examples.  Since the authors of the dataset are from the University of Amsterdam, we refer to the dataset as the UA data. The second data was recorded at MIT Place-Lab by Tapia {\it et al.} \cite{mitdata2003,mitdata2004}. The data was extracted from two single-person apartments (subject one and subject two). The apartments were fitted with 76 and 70 sensors for subject one and subject two, respectively, and the data was collected for two weeks. The data has 14 example sequences of length 1440 time steps. Annotated activities are categorized into eight high level activities such as $employmentRelated$, $personalNeeds$, $domesticWork$, $educational$, $entertainment$, {\it etc.}. We refer to this dataset as MIT data. The third is an relational activity recognition data provided by Gutmann and Kersting ~\cite{relTransorm} of Katholieke University, Leuven.  The data has been collected from a kitchen environment with 25 sensors/RFID attached to objects. There are 19 activities such as $boilWater$, $makeTea$, $toastBread$, $pourMilk$, $eat$, $obtainNewspaper$, {\it etc.}. The data contains 20 sequences, each of approximate length 250 time steps. Unlike the other two datasets, this data has at-most one sensor on at any given point of time. We refer to the data as KU data. We now start with the experiments and results of our first contribution, the StructHKL framework to learn simple conjunctions ($\mathcal SC$) for sequence labeling (for example, $microwave$ and $platesCupboard$ firing at the same time step).

In this set of experiments, since we are interested in discovering the input structure at a single sequence position, UA data and MIT data fit our bill. Since KU data doesn't have multiple sensors fired at a single time step, there is no $\mathcal SC$ other than the trivial one (one predicate) that can be discovered, and hence is not suitable to evaluate the effectiveness of StructHKL.

We performed our experiments in a 4 fold cross validation set up. In each cross validation experiment, we used 25\% of data for training\footnote{\scriptsize In real world applications, annotating the activities is a manual and expensive task. Hence, it is desirable that approaches should be able to build models without using large amount of data.} and the rest for testing and report all accuracies by the average across the four folds. We report both micro-average and macro-average prediction accuracies. The micro-average accuracy is referred to as time-slice accuracy by~\cite{kasteren08}, and is the average of per-class accuracies, weighted by the number of instances of the class. Macro-average accuracy, referred to as class accuracy by~\cite{kasteren08}, is simply the average of the per-class accuracies. Micro-averaged accuracy is typically used as the performance evaluation measure and therefore, our objectives are derived for improving micro-average accuracies\footnote{\scriptsize In our approaches, we optimize on micro-average accuracy and hence our results are good in micro-average accuracy while giving comparable results for macro-average accuracy.}. However, in data that is biased towards some classes, too bad a macro-average accuracy is an indicator of a bad prediction model. 
In the following paragraphs, we compare our approach with other approaches that gave comparable results.

For the UA data, we compared our results with eight other approaches: (a) standard HMM~\cite{hmm}, (b) Branch and Bound structure learning assisted HMM model construction (B\&B HMM), where the rules learned by Aleph~\cite{aleph} (an ILP system which learns definite rules from examples) for each activity determine the HMM emission structure, (c) greedy feature induction assisted HMM approach (Greedy FIHMM)~\cite{naveenActivity}, (d) StructSVM approach~\cite{TsochantaridisSVMStructured}, (e) Conditional Random Field (CRF)~\cite{CRFLafferty}, (f) Conditional Random Field with Feature Induction (FICRF)~\cite{CRFMcCallum,McCallumMALLET}, (g) RELHKL (without considering transitions)~\cite{ganeshRELHKL} and (h) RELHKL + StructSVM. While standard approaches such as HMM, CRF and structSVM use basic features (binary sensor values) as emission features, feature induction approaches such as Greedy FIHMM and FICRF use conjunctions of basic features as emission features. In contrast to greedy feature induction approaches,
RELHKL, and StructHKL find the feature conjunctions efficiently and optimally. While RELHKL, without the transition features, does not consider the structure in output space, RELHKL + StructSVM solves the problem in two steps. In the first step, RELHKL (without considering transitions) is employed to learn rules for each label. In the second step, the rules learned in the first step are fed as features into the StructSVM algorithm to get the final model. In contrast, StructHKL does the classification in structured output space (rules and parameters are learned simultaneously for structured output classification) and performs better. Our model parameter tuning for all the experiments were not performed in the best desired way, since the huge problem space limits the scalability of approaches tried. However, we have tuned the parameters in the best possible way, given the constraints and resources we had\footnote{\scriptsize For making our approaches scalable on such large data, we suggest, (i) parallelization using hadoop, (ii) using heuristics in our computations, and (iii) allowing larger epsilon values (amount by which the output from certain procedures can deviate from the optimum). We leave these for future work.}. The results are summarised in Table 1. We observed that the proposed StructHKL approach outperforms all the other approaches in micro-averaged accuracy (which is our objective). While our macro average accuracy is comparable to FICRF, StructSVM, RELHKL + StructSVM and outperforms others, our standard deviation is much less, which reflects its consistency.

\begin{table}[]
% \scriptsize
% \parbox{.48\linewidth}{
\centering
%\tbl{Micro and macro average accuracies in percentage of different approaches on UA data.\label{tableRawUA}}{%
\label{tableRawUA}{
  \begin{tabular}[m]{lcc}
  \hline
  \texttt{} & \texttt{Micro avg.} & \texttt{Macro avg.}\\
  \hline
  \texttt{Std. HMM} & 25.40 ($\pm$18.55) & 21.75 ($\pm$12.12) \\
  \texttt{B\&B HMM} & 29.54 ($\pm$20.70) & 16.39 ($\pm$02.74) \\
  \texttt{Greedy FIHMM} & 58.08 ($\pm$10.14)  & 26.84 ($\pm$04.41) \\
  \texttt{StructSVM} & 58.02 ($\pm$11.87)  & 35.00 ($\pm$05.24) \\
  \texttt{CRF} & 48.49 ($\pm$05.02)  & 20.65 ($\pm$04.82) \\
  \texttt{FICRF} & 59.52 ($\pm$11.76)  & 33.60 ($\pm$07.38) \\
  \texttt{RELHKL} & 46.28 ($\pm$11.44)  & 23.11 ($\pm$07.46) \\
  \texttt{RELHKL+StructSVM} & 55.74 ($\pm$10.88)  & 38.56 ($\pm$10.68) \\
   \texttt{StructHKL} & 63.96 ($\pm$05.74)  & 32.01 ($\pm$03.04) \\
  \hline
  \end{tabular}}
%  \caption{Micro and macro average accuracies in percentage of different approaches on UA data.}
%%  }
% \label{tableRawUA}
% }
\end{table}
% \hspace{0.4cm}
% \parbox{.48\linewidth}{
\begin{table}[]
\centering
%\tbl{Micro and macro average accuracies in percentage of different approaches on MIT data\label{tableMIT}}{%
\label{tableMIT} {
  \begin{tabular}[m]{l|lcc}
  \hline
  \texttt{} && \texttt{Micro avg.} & \texttt{Macro avg.}\\
  \hline
  \multirow{4}{*}{\rotatebox{90}{Subj 1}} &\texttt{StructSVM} & 75.03 ($\pm$04.51)  & 26.99 ($\pm$07.73) \\
  &\texttt{CRF} & 65.54 ($\pm$06.80)  & 31.19 ($\pm$07.39) \\
  &\texttt{FICRF} & 68.52 ($\pm$07.19)  & 29.77 ($\pm$03.59) \\
  & \texttt{StructHKL} & 82.88 ($\pm$0.43)  & 28.92 ($\pm$01.53) \\
  \hline
  \multirow{4}{*}{\rotatebox{90}{Subj 2}}&\texttt{StructSVM} & 63.49 ($\pm$02.75)  & 25.33 ($\pm$05.8) \\
  &\texttt{CRF} & 50.23 ($\pm$06.80)  & 27.42 ($\pm$07.65) \\
  &\texttt{FICRF} & 51.86 ($\pm$07.35)  & 26.11 ($\pm$05.89) \\
  & \texttt{StructHKL} & 67.16 ($\pm$08.64)  & 24.32 ($\pm$02.12) \\
  \hline
  \end{tabular}}
%\caption{Micro and macro average accuracies in percentage of different approaches on MIT data}
%}
%%.}
% \label{tableMIT}
%%  }
\end{table}

In our experiments on MIT dataset, we observed that the performance of standard HMM, B\&B structure learning assisted HMM, and RELHKL without transition features was poor and the greedy feature induction assisted HMM did not converge at all. Therefore, we compare our results with (a) StructSVM approach~\cite{TsochantaridisSVMStructured}, (b) Conditional Random Field (CRF)~\cite{CRFLafferty}, and (c) Conditional Random Field with Feature Induction (FICRF)~\cite{CRFMcCallum,McCallumMALLET}. The results are summarised in Table 2. Our results show that StructHKL performs better than other approaches in micro-averaged accuracy for both subject one and two, while maintaining comparable macro-averaged class accuracies. Our approach shows less standard deviation in the data of subject one while showing comparable standard deviation in the data of subject two.

We computed statistical significance tests for micro-average using Wilcoxon Signed Rank Test \cite{wilcoxon}, where we paired the results from each cross validation experiment of our approach with the corresponding experiment of a competitor approach. Our statistical significance tests for micro-average indicate a 0.01 level of significance over all other approaches we compared against for both the UA dataset and the MIT dataset.

In a setting with {\footnotesize $n$} labels and {\footnotesize $N$} basic inputs, an exhaustive search for optimum features needs evaluation at {\footnotesize $n\times 2^N$} nodes (conjunctions). This amounts to 131072 nodes in UA data and to the order of $10^{22}$ in MIT data, which is computationally infeasible. In contrast, due to the active-set algorithm and sufficiency condition check, our approach explores only a few thousand nodes and converges in 30 hours approximately. In our experiments we have observed that traditional sequence labeling algorithms such as HMM and CRF, greedy feature induction approaches such as FIHMM and FICRF take a few minutes for training. StructSVM's running time ranges between a few minutes to a few days, depending on the regularisation parameter used. The training times of different approaches for UA data are shown in table \ref{tableRawUARunTime} Since all approaches use dynamic programming for prediction, time for inference is similar.

\begin{table}[]
% \scriptsize
\centering
%\tbl{Training time of different approaches on UA data.\label{tableRawUARunTime}}{%
\label{tableRawUARunTime} {
  \begin{tabular}[m]{lc}
  \hline
  \texttt{} & \texttt{Training time}\\
  \hline
  \texttt{Std. HMM} & 1 second \\
  \texttt{B\&B HMM} & 1 second \\
  \texttt{Greedy FIHMM} & 2.73 minutes \\
  \texttt{StructSVM} & 8.12 minutes\\
  \texttt{CRF} & 1.95 minutes\\
  \texttt{FICRF} & 2.57 minutes\\
  \texttt{RELHKL} & 104 minutes\\
  \texttt{RELHKL+StructSVM} & 120 minutes\\
   \texttt{StructHKL} & 31 hours\\
  \hline
  \end{tabular}
%  \caption{Training time of different approaches on UA data.}
% \label{tableRawUARunTime}
 }
\end{table}

The StructHKL approach discovered rules such as :

\vspace{-0.1cm}
{\footnotesize  
\begin{itemize}
 \item[]$usingToilet(T) \leftarrow bathroomDoor(T) \wedge  toiletFlush(T)$,
  \item[]$sleeping(T) \leftarrow bedroomDoor(T) \wedge toiletDoor(T) \wedge bathroomDoor(T)$,
  \item[]$preparingDinner(T) \leftarrow groceriesCupboard(T)$
\end{itemize}
}

\vspace{-0.1cm}
The conjunction {\footnotesize $bathroomDoor(T) \wedge  toiletFlush(T)$} strongly indicates that the activity is {\footnotesize $usingToilet$}  while {\footnotesize $groceriesCupboard$} indicates a higher chance of {\footnotesize $preparingDinner$}. Similarly {\footnotesize $bedroomDoor(T) \wedge toiletDoor(T) \wedge bathroomDoor(T)$} increases the chance of predicting {\footnotesize $sleeping$} as the activity. This is reasonable, as people access these doors during night before going to sleep, and the sensors at {\footnotesize $bedroomDoor$, $toiletDoor$}, and {\footnotesize $bathroomDoor$} fire once, when the person accesses these doors and goes to {\em off-mode} while s/he is sleeping. However, since the conjunction just before sleep gives a higher weight to the activity {\footnotesize $sleeping$}, the weight gets accrued and gets combined with transition weights to accurately predict the activity as {\footnotesize $sleeping$}. We now discuss our experiments on learning complex relational features derived from relative sequence positions (for example, $platesCupboard$ firing 1 minute after $microwave$ became on). An example of such a feature discovered in our experiments (this is an $\mathcal{AF}$  discovered by Warmr) is as follows: 

{\footnotesize
$preparingDinner(T1) \leftarrow pansCupboard(T1) \wedge \\ \text{\scriptsize{.}} \qquad \qquad \qquad \qquad \qquad \qquad  nextRelPosWindowNear(T1, T2) \wedge cupsCupboard(T2)$
}

This relational features suggests that {\footnotesize $preparingDinner$} is an activity which involves sensors at different temporal instances. Using this enumerated feature in the model enhances its expressivity  than a simple conjunction of features at the same time instance as discussed in the previous sections.

For evaluation of our two approaches for learning $\mathcal{CF}$, {\it viz.} enumerating $\mathcal{AF}$ (enum$\mathcal{AF}$) and relational subsequence kernels (subseqSVM), we use the UA data and KU data. We perform a 4 fold cross validation with 25\% train and 75\% test sizes for UA data. For KU data, we perform experiments in a leave one out cross validation set up\footnote{\scriptsize Since the number of data points in KU data is fewer, in each cross validation experiment, we train on all example sequences except one and test on the left out example sequence}. Since we are exploring a huge space (relational features spanning over multiple inputs and multiple sequence positions), our experiments did not scale for MIT dataset, which has more than 70 individual inputs. We therefore, present results from experiments performed on UA data and KU data.

We have compared our approaches  with TildeCRF~\cite{tildeCRF} and StructSVM~\cite{TsochantaridisSVMStructured}. TildeCRF is the state-of-the-art ILP approach to learn relational features for sequence labeling, and operates in the same feature space that we are interested in, while we treat StructSVM (only basic inputs as features) as the baseline for this experiment. We also report StructHKL results for UA data for the ease of reference. Since KU data doesn't have multiple inputs at each sequence position, there is no structure that can be discovered by StructHKL and therefore, only StructSVM results are reported.

The comparison of results on the UA data is outlined in Table~\ref{tableMLCvsTildeCRFUAdata}. Results show that our approaches to learn complex features for sequence labeling namely, enum$\mathcal{AF}$ and SubseqSVM, performed better than the base line approach (StructSVM) and the state-of-the-art approach (TildeCRF). Although enum$\mathcal{AF}$ optimally finds $\mathcal{CF}$s as conjunctions of (selectively enumerated) $\mathcal{AF}$s, the step for selectively enumerating $\mathcal{AF}$s is based on heuristics. In contrast, SubseqSVM works on a convex formulation and learns an optimal model. This explains the difference in performance of our two approaches.

The comparison of results for KU data is outlined in Table~\ref{tableMLCvsTildeCRFKUata}. As a single sequence step in this data has only one input feature, the feature space is not rich enough to evaluate the efficiency of our approaches. For this reason, the performance of our approaches is inferior to the baseline and the state-of-the-art.
The baseline reported the best performance. While the performance of SubseqSVM approach is slightly inferior to the baseline and the state-of-the-art, enum$\mathcal{AF}$ performed badly in this data.

In the case of the UA data, both our approaches (enum$\mathcal{AF}$ and SubseqSVM) longer to train the model, than the competitors. Inference with SubseqSVM takes, on average, 6 hours for UA data, Whereas, other approaches take only a few seconds for inference. The difference is due to kernel computation. Our statistical significance test for micro-average using Wilcoxon Signed Rank Test \cite{wilcoxon} indicates a 0.01 level of significance with SubseqSVM over other approaches on UA data.

\begin{table}
% \scriptsize
% \parbox{.40\linewidth}{
\centering
%\tbl{Micro and macro average accuracies in percentage of different approaches on UA data.\label{tableMLCvsTildeCRFUAdata}}{%
\label{tableMLCvsTildeCRFUAdata} {
  \begin{tabular}[m]{lcc}
  \hline
  & \texttt{Micro avg.} & \texttt{Macro avg.}\\
  \hline
  \texttt{tildeCRF} &  56.22($\pm$12.08)  & 35.36 ($\pm$06.55) \\
  \texttt{StructSVM}  &  58.02 ($\pm$11.87)  & 35.00 ($\pm$05.24) \\
\texttt{StructHKL} & 63.96 ($\pm$05.74)  & 32.01 ($\pm$03.04) \\
  \texttt{enum$\mathcal{AF}$} & 60.36 ($\pm$06.99)  & 30.39 ($\pm$04.31) \\
  \texttt{SubseqSVM} &  65.25($\pm$04.81)  & 29.34 ($\pm$02.78) \\
   \hline
  \end{tabular}}
% \captionsetup{width=0.4\textwidth}
%\caption{Micro and macro average accuracies in percentage of different approaches on UA data.}
% \label{tableMLCvsTildeCRFUAdata}
\end{table}
% \hspace{0.85cm}
% \hfill
% \parbox{.48\linewidth}{
\begin{table}
\centering
%\tbl{Micro and macro average accuracies in percentage of different approaches on KU data.\label{tableMLCvsTildeCRFKUata}}{%
\label{tableMLCvsTildeCRFKUata} {
  \begin{tabular}[m]{lcc}
  \hline
  & \texttt{Micro avg.} & \texttt{Macro avg.}\\
  \hline
  \texttt{tildeCRF} & 66.04 ($\pm13.50$)  & 84.01 ($\pm$08.76) \\
  \texttt{StructSVM}  & 66.35 ($\pm$17.16)  & 66.64 ($\pm$16.04) \\
  \texttt{enum$\mathcal{AF}$} & 33.24 ($\pm15.72$)  &  23.02 ($\pm$11.13) \\
  \texttt{SubseqSVM} & 64.66 ($\pm$08.42)  & 63.08 ($\pm$07.05) \\
  \hline
  \end{tabular}}
% \captionsetup{width=.4\textwidth}
%\caption{Micro and macro average accuracies in percentage of different approaches on KU data.}
% \label{tableMLCvsTildeCRFKUata}
% }
\end{table}

\begin{table}[]
% \scriptsize
\centering
%\tbl{Training time of different relational approaches on UA data.\label{tableRawUARunTime2}}{%
\label{tableRawUARunTime2} {
  \begin{tabular}[m]{lc}
  \hline
  \texttt{} & \texttt{Training time}\\
  \hline
  \texttt{tildeCRF} & 2.3 hours \\
  \texttt{StructSVM}  & 8.12 minutes \\
  \texttt{enum$\mathcal{AF}$} & 21.28 hours \\
  \texttt{SubseqSVM} & 18.21 hours \\
  \hline
  \end{tabular}
%  \caption{Training time of different relational approaches on UA data.}
% \label{tableRawUARunTime2}
 }
\end{table}

\section{Conclusion}
\label{sec:conclusion}

Recent work has shown the importance of learning input structure, in the form of relational features, for sequence labeling problems. Most existing feature learning approaches employ greedy search techniques to discover relational features. In this work, we studied the possibility of optimally learning relational observation features while preserving all transition features. First, we categorized relational features and identified interesting feature categories. We have proposed a hierarchical kernel learning approach for structured output spaces (StructHKL), which can optimally discover features derived from individual inputs at a single sequence step. However, StructHKL has limitations in discovering complex features that are derived from basic inputs at relative positions. To this end, we identified a class of features called {\em composite features}, that are conjunctions of features belonging to a simpler category called {\em absolute features}. We proposed and developed two strategies to learn optimal 
composite features. One, to selectively enumerate absolute features and employ StructHKL to learn their conjunctions. Two, to incorporate relational subsequence kernels, that implicitly capture the information about all possible composite features, without explicitly enumerating them. We have evaluated our approaches on publicly available activity recognition datasets. From our results, we observe that for every choice of feature language (ranging from propositional to composite features), our HKL and kernel based feature construction approaches outperform the baselines on micro average accuracies, while yielding slightly worse macro average accuracies but with reduced standard deviations. 

\bibliographystyle{plain}
\bibliography{journal}
% Sample .bib file with references that match those in
% the 'Specifications Document (V1.5)' as well containing
% 'legacy' bibs and bibs with 'alternate codings'.
% Gerry Murray - March 2012

% History dates
%\received{February 2007}{March 2009}{June 2009}

% Electronic Appendix
%\elecappendix

%\medskip

\end{document}